# Hybridization of K-means with improved firefly algorithm for automatic clustering in high dimension

Afroj Alam [1]
Department of Computer Application
Integral University
Lucknow, India
alamafroj@student.iul.ac.in

Mohd Muqeem 2
Department of Computer Application
Integral University
Lucknow, India
muqeem@iul.ac.in

*Abstract*— K-means Clustering is the most well-known partitioning algorithm among all clustering, by which we can partition the data objects very easily in to more than one clusters. However, for K-means to choose an appropriate number of clusters without any prior domain knowledge about the dataset is challenging, especially in high-dimensional data objects. Hence, we have implemented the Silhouette and Elbow methods with PCA to find an optimal number of clusters. Also, previously, so many meta-heuristic swarm intelligence algorithms inspired by nature have been employed to handle the automatic data clustering problem. Firefly is efficient and robust for automatic clustering. However, in the Firefly algorithm, the entire population is automatically subdivided into sub-populations that decrease the convergence rate speed and trapping to local minima in high-dimensional optimization problems. Thus, our study proposed an enhanced firefly, i.e., a hybridized K-means with an ODFA model for automatic clustering. The experimental part shows output and graphs of the Silhouette and Elbow methods as well as the Firefly algorithm.

I. INTRODUCTION

Clustering is an important, influential, and unsupervised machine learning technique for natural grouping the objects into groups according to their similarity [1]. Objects will be segmented into different clusters based on some cluster validity index (CVI) metrics. This CVI shows the relation between cluster cohesion i.e. intra-cluster distance within group and cluster separation i.e. inter-cluster distance between groups. This process of partitioning the objects is called clustering. These objects do not have any external information, such as class labels; that is why clustering is an instance of unsupervised machine learning classification techniques [2]. Data clustering is very useful for partitioning the data so that we can make exploratory data analysis to find the hidden patterns and valuable information from high-dimensional data [3].

Nowadays, the clustering technique is being applied in diverse field of real-world problems, e.g. in engineering, wireless sensor networks, mobile networks, medical science, computer science, biological science, earth science, economics, and bioinformatics, yield-marketing, image analysis, web mining, spatial database analysis, insurance, statistical data analysis, fraud detection, libraries (book ordering), loan approval, community detection (e.g., LinkedIn, Facebook, Twitter, etc.), pattern recognition, data compression, classification of plants and animals, improving decision-making in business intelligence, etc.[4-7]. At the starting clustering approach were used mainly in two domains i.e. anthropology and psychology, which were the two social science areas. Further with advancement; it's implemented also in trait theory. After that, it extended to other new study directions with major impacts like machine learning and data science. As a result, data clustering is a significant and hot topic in many domains. E.g. in artificial intelligence as well as in data mining [8-9].

Researchers have proposed several partitioning-based heuristic algorithms from last 2-3 decades to solve the clustering problems. Among all partitioning clustering algorithms, the K-Means method appeared as a most prominent and widely used and powerful algorithm for selecting the optimum number of output clusters in a dataset. Because it is straightforward to implement and flexible, its run time complexity is less than other clustering algorithms; it is also rated under the top 10 data widely used in data mining [10]. K-means clustering is a non-deterministic method for dividing n data points into K non-overlapping clusters in an N-dimensional space. There should be not any fuzzy data points in any cluster. However, there are a lot of major challenges with K-means clustering, such as it highly relies on a predefined appropriate number of clusters K for dataset having high dimensional. This leads the algorithm to get trap to local optima. Defining and initializing the appropriate number of clusters and their corresponding centroids had made the limitation of performance and their accuracy of the quality of good clusters [11]. Also, K-means clustering is NP-hard in the high dimensional dataset; there is no any widely accepted theory of K-means for all types of datasets. Above drawbacks prompted data mining experts to develop new methods to address them and come up with other effective means to enhance the K-Means? To address the shortcomings as mentioned earlier and tackle more sophisticated and high-dimensional data clustering challenges, the data mining researcher redefines the K-means, which is hybridized with meta-heuristic optimization methods such as swarm intelligence algorithms [12-13].

There are a lot of stochastic meta-heuristic multi-model evolutionary algorithms which is inspired by nature. These multi-model algorithms have been used to solve the challenge of data clustering, which are differential evolution (DE), Tabu search algorithm, and genetic algorithm (GA). At the same time, there are several swarm intelligences based techniques such as symbiotic organisms search (SOS), PSO, ant colony optimization (ACO), invasive weed optimization (IWO), firefly algorithm (FA), Cuckoo algorithm(CA), ABC, teaching learning-based optimization (TLBO) etc. all these algorithms integrated with clustering to solve the clustering problems. For Clustering analysis swarm intelligence and evolutionary algorithms both act as an optimization catalyst in global search space to maximize the cohesive process inside the clusters and maximize the adhesive process for separation the clusters. The above two stochastic meta-heuristic methods are far better than traditional clustering in high-dimensional data clustering problems. These nature-inspired stochastic optimization algorithms have been widely used in different fields for solving a real-life optimization problem. Like GA is used to solve high-dimensional clustering problems, location problems, and flow shop problems. Differential Evolutionary (DE) algorithms have been widely used to answer computer science problems and engineering optimization complex problems. The PSO is also successful in several high-profile applications in terms of high accuracy and good convergent speed [14-15]. It has been observed that using NIC not only cause's automatic clustering problems solves, it also improves the accuracy, efficiency, and robustness of the algorithms if we hybridize them with another traditional algorithm [16-17].

However, for the majority of above stochastic meta-heuristic optimization algorithms we have to prior specify the value of K for number of clusters as well as structure of data and their parameters in K-means. Unfortunately, to find the optimal value of K in advance for KM clustering is a challenging task in high dimensional dataset. In many successful applications, the FA is another well-known popular meta-heuristic optimization technique that has been utilized to solve real life complex application, challenges of data clustering, and scheduling problems which is not related to parallel machine execution [18]. Hence, our research idea proposed a hybridized K-means clustering with an enhanced FA method; this enhanced FA not required and prior input. It will automatically calculate K i.e. number of clusters in this manner automatically. This automatic calculated value of K by FA will help the k-means algorithm to converge to the global optima and generates a quality of clusters; this is our proposed methodology.

Further the outline of our paper of remaining part is as given below: In section II we have described briefly and scientifically the exhaustive literature review on the k-means clustering algorithms, metaheuristic techniques for global optimization, and procedure for automatic clustering. Section III elaborates on the K-means clustering. Section IV elaborates on the Firefly Algorithms and its variants design concept. Section V gives a detailed description of the proposed algorithm. In section VI, Experimental results are done, and Section VII concludes the paper.

## II. LITERATURE REVIEW

The researcher Indra kumari et al. [19] in his paper proposed an improved K-means that will not generate any empty clusters in K-means. They also showed in his work that the upgraded k-means cluster has a lower running time complexity than a conventional k-means cluster. They used the concept of the backtracking method algorithm, by which the algorithm takes the help of previously calculated data for calculating the new centroids. For this algorithm, the future challenges and work is space minimization of this algorithm.

Islam et al. [20] has proposed a more efficient nature-inspired algorithm, i.e., Genclust, which combines the capacity of a genetic algorithm to the conglomerate different solutions with the exploitation of a hill-climber for the entire search space. The researcher has improved the traditional genetic searching approach by using the concept of hill-climbing concepts of K-means, which is faster than the traditional clustering and gives the results of higher quality clusters, and also reduces the computational resources. A researcher has also tested their algorithm on different datasets.

According to the current researcher Alam et al. [21], K-means clustering is more general and easy for the prediction of disease in the field of health care. This algorithm is also prominent for the detection of fraud in the bank, detection of crime, yield management, and profitability analysis. This algorithm also applies to international super-market for predicting frequent item sets for customer attraction. We need some suitable data mining techniques like K-means, which hybridized with metaheuristic algorithm to improve the accuracy of real-life NP-hard optimization problems.

Shi et al. [22] has improved the effectiveness of data analysis in his proposed hybridized algorithm over the classical K-means algorithm, which was easily convergent to the local optima by the initial selection of cluster centroids. The proposed hybridized algorithm is a genetic algorithm and the K-means. In this paper genetic algorithm is used for cleaning and reducing the dimensionality of the datasets, then uses the K-means to optimize the selection of cluster centroids. Three different approaches they have tested are: standard K-means, K-means hybridized with GA, and population-based GA with K-means clustering. Objective of all the three approaches is to automatic selection of initial cluster centroids. The proposed method gives more accuracy.

The Analyst Hrosik et al. proposed a hybridized approach for the division of brain picture for recognizing diverse essential tumors. The proposed idea placed a strong emphasis on tracking various primary brain tumors: glioma, metastatic bronchogenic carcinoma, metastatic adenocarcinoma, and sarcoma. This method is analyzed with standard benchmark digital images, and it improves the quality of cluster results compared to other simple K-means clustering methods [23].

The researcher has proposed an image analysis for MRI pattern recognition in brain tumor with K-means as unsupervised machine learning and nature-inspired-based PSO and FA. They improved the image division using the fitness function of Swarm-Based PSO. The accuracy of segregation of image is improved with the help of KM and PSO. The Comparative studies have shown that the combined k-means with FA exhibited high accuracy and precision in detecting brain tumor Region-of-Interest [24].

## III. K-MEANS CLUSTERING

The K-means clustering algorithm is an unsupervised machine learning technique that divides data into K clusters based on inter-cluster and intra-cluster distances. The objective of this algorithm is to minimize the intra-cluster, i.e., Euclidean distance should be minimized to get good quality clusters. This technique assumes that a data object belongs to one of two clusters: one or none. It has been one of the most extensively utilized clustering algorithms for tackling numerous real-life situations due to its simplicity, ease, and linear time running complexity [25]. We can explore this algorithm in such a way that we have T sets of data and we have to segregate it into K sets with non-overlapping. $S=\{s1_1, s2_2, s_3,.....s_k\}, s_j \neq \Phi, j=1,2,...,k$ where $T= U_1^k s_j$ ; $s_j \cap s_i = \Phi, j,i=1............k$ and $j \neq i$. During the partitioning process, we have to optimize the fitness function, i.e., Minimize intra-cluster distances between objects within the cluster while maximizing inter-cluster distances between objects between clusters [13], and this is the objective function of K-means clustering, which we have to achieve for good quality of clusters. The objective function can be defined mathematically as follows.

The given dataset $T=\{t_1, t_2, t_3,.............,t_n\}$ has a d-dimension. T is being isolated for k sets as given below,

$$P(s_k) = \sum_{s_i \in t_k} \|s_i - \mu_k\|^2 \qquad (1)$$

so that the square errors sum objective function should decline across all k clusters. i.e., minimize the given equation

$$P(t) = \sum_{k=1}^{K} \sum_{s_i \in t_k} \|s_i - \mu_k\|^2 \qquad (2)$$

The main focus of automatic clustering is to identify the optimal value of k, and the distribution of data objects should be correct in all the clusters. As a result, in automatic

Table [1]. Comparative analysis of FA for automatic clustering

| References | Variants of FA | Application | Dataset | Comparison with | Result |
|---|---|---|---|---|---|
| [26] | Opposition based FA | Enhancement of Medical Image | Medical data | FFA, AMFF (Adaptive moment fireflies) | Enhancement in contrast |
| [27] | FA | Prediction of Disease and classification of data | IRIS, Glass, Wine, Heart, Breast-cancer | PSO, GA, ABC | Accuracy of classification and disease prediction is improved |
| [28] | Hybridized FA | Identification Myocardial infarction on text data | Diabetes data | Traditional FA, Bat algorithms | Accuracy and sensitivity are enhanced in comparison to others |
| [29] | Binary FA | DNA-binding proteins | Protein Data Bank | Statistical FA | Binary FA outperforms than Statistical FA |
| [30] | Hybrid FA | Image segmentation | MRI scan, Rice, Lena and satellite image | IFCM (Improved Fuzzy C-Means), FCM (Fuzzy C-Means) | IFCMFA outperforms IFCA in terms of performance |
| [31] | Moth moth-flame optimization | facial expression recognition | CK+ JAFFE, and MMI | GA, PSO etc | Compared to other, moth-flame optimization outperforms. |
| [32] | Chaotic FA | MR brain image segmentation | IBSR Dataset | Fuzzy C-Means | Accuracy of segmentation Rate improved |
| [33] | FAFCM | real world complex datasets clustering | MRI Image data | FCM and PSO | FCM is outperformed by FACFM. |
| [34] | FA | Classifiers and feature extractors | BCI competition dataset | GA | Superior performance than existing algorithm |

clustering, we must optimize the number of choices used to allocate R data objects to K groups.

The optimal value of K is being determined mathematically in the search space is as below:

$$D(R) = \sum_{K=1}^{r} P(R, K) \qquad (3)$$

Here is P is the search space. To find an optimal solution this is an NP-hard issue with K>3. The computational time of a task with a high-dimensional and huge dataset is very high. It is challenging to perform automatic clustering for such datasets without prior knowledge of the data items' characteristics. Hence creating an appropriate no of clusters and distributing the objects into their corresponding clusters is computationally exhaustive and time-consuming in traditional K-means clustering [35].

A. *Principal Component analysis (PCA)*

Recent methodology of KM for data having higher dimension use deep learning method to select attributes which are highly informative of a given data set. K-means clustering will then be applied to the reduced data set. Boutsidis et al. [36] proposed an algorithm that helps to decreases dimension to t from q by selecting a small subset of m rows from the data matrix D ε R $^{qXn}$ where t<q. Then, on data matrix D, apply the KM approach to cluster the data objects into K groups. Consideration of variance-covariance for good clusters in large dataset, the authors executed PCA which will include only informative attribute [37].

With the help of PCA we can filter out irrelevant features, which reduce the training time as well as cost, and improve the performance of given model . After PCA implementation the dataset will be passed to the K-means clustering to reduce the outlier's data from the clusters. [38].

PCA is a technique for dimension reduction that filters out uncorrelated variables from high dimensional data which do not adequately explain the original variables.

PCA is a technique for dimension reduction that filters out uncorrelated variables from high dimensional data which do not adequately explain the original variables.

Suppose the matrix $Y^T=[Y_1,Y_2,Y_3,\ldots, Y_p]$ and the their matrix $\Sigma$ with eigenvalues $\lambda_1 >= \ldots \lambda_p >=0$ is covariance

$X_1 = a_1^T Y = a_{11}Y_1 + a_{21}Y_2 + \ldots a_{p1}Y_p$

$X_2 = a_2^T Y = a_{12}Y_1 + a_{22}Y_2 + \ldots a_{p2}Y_p$

$X_p = a_p^T Y = a_{1p}Y_1 + a_{2p}Y_2 + \ldots a_{pp}Y_p$

All of these equations can be replaced by $X_i$, the $i^{th}$ principal component in the given below equation.

$$Variance(X_i) = a_i^T \sum a_i = e_i^T \sum e_i; \quad (4)$$

$$Co-Variance(X_i, X_k) = a_i^T \sum a_k = e_i^T \sum e_k; \quad (5)$$

The linear combination with the greatest variance is the first principal component. In other words, it maximizes Var $(Y^1)$ = $a_1^T \Sigma a1$. It is Var(Y1) = $a_1^T \Sigma a1$ can clearly be increased by multiplying any $a_1$ by some constant. Hence first and second principal component in (7) and (8).

$$PC1 = \begin{cases} Maximize: Variance(X_1) \\ Subject\ to\ a_1^T a_1 = 1 \end{cases} \quad (6)$$

$$PC2 = f(x) = \begin{cases} Maximize: Variance(X_2) \\ Subject\ to: a_2^T a_2 \\ CoVariance\ (X_1, X_2) \end{cases} \quad (7)$$

In the next stage of our algorithm, this PC1 and PC2 will be used as input for our KM method. Below is the python algorithm for PCA.

```
import the iris dataset
import PCA from sklearn
include numpy as np1
def plt_clustered_out(X, centers):
    pca = PCA(2).fit_transform(np1.concatenate([X, centers], axis=0))

true_centers = np.array([np1.mean(iris_data[..], axis=(0)) for label in set(iris_labels)])
plt_clustered_out(iris_data, true_centers
```

Clustering graph of clustering of iris data after PCA implementation is in experimental part Fig. 1.

*B. Various approaches to identify value of k in extended K-means:*

Without uses of deep learning, prior finding the optimal value of K in KM is a challenging tas., hence KM will converge to local optimal solution. Different researcher used different approaches. Some of them explained given below:

I. Silhouette method:

Silhouette method is a clustering validation index tool for optimal cluster numbers. This method provides the graphical representation of intra as well as inter cluster distances within its clusters and data points. The Individual silhouette coefficient index sc(k) value for a data point i is defined as a ratio scale data which is explained in equation (7) and in Fig (2).

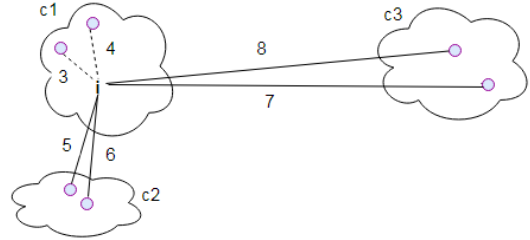

Fig. 2. Example of the Silhouette coefficient.

$$sc(i) = \frac{q(k) - p(k)}{\max\{p(k), q(k)\}} \quad (8)$$

p(k) = mean value of dissimilarity of data point k to all other data points of c1.

p(k)=(3+4)/2=3.5

q(k) = minimum average dissimilarity between data point i to all other clusters $c_2$ and $c_3$

q(k)=min((8+7)/2,(5+6)/2)=5.5

Output of equation (3) is in between -1 and +1 i.e. -1 <= sc(k) <=+1 .

+1 of s(i) indicates that data points are more dense and data are correctly clustered. If its value is near to -1 means bad clustering, then data point k would be more appropriate, if it would be clustered in its neighbouring cluster. 0 shows intermediate cluster. Equation (7) Calculates the intra-inter Silhouettes value for a given data point.
Hence, larger silhouette are well clustered as compared to small silhouette statistic value between clusters. The average value of sc(k) shows how data are tightly dense in the cluster and also it measures how entire data has been clustered correctly by given below equation [39].

$$\overline{SC} = \frac{1}{n}\sum_{k=1}^{n} SC(k) \quad (9)$$

algorithm is given below for **Silhouette** index
```
m = KMeans(random_state=1) //m is model
v = KElbowVisualizer(m, k=(2,10), metric='silhouette')
v.fit(X_numerics) // v is visualizer
v.show()
plt.show()
```

Experimental results for above method is given in Fig. 3. In experimental part. Silhouette score method indicates the best options would be 5 or 6 clusters in Fig.3.

Using Silhouette we can also check the quality of clusters. Code is given below and experimental analysis results is in experiment part in Fig. 4.

```
from yellowbrick.cluster import SilhouetteVisualizer
m = KMeans(nc=5, rs=0)
v= SV(model, colors='greenbrick')
v.fit(X_numerics)
visualizer.show()
plt.show()
```
here m is model , v is visualizer, SV is SilhouetteVisualizer
II. Elbow Method

It is a graphical and old approach for finding the optimal number of K in K-means algorithm. It uses the concepts of sum of squares error (SSE) as an objective function for clustering validity as well as cluster quality. SSE is defined as follows.

$$SSE = \sum_{k=1}^{K} \sum_{x_i \in s_k} \|X_i - C_k\|_2^2 \quad (10)$$

with K= number of clusters, Sk is the k-th cluster, xi is the element of kth cluster, Ck is the centeroid of the cluster Sk, ‖ · ‖ is the Euclidean distance between two data patterns.
We have to plot the graph of line chart between SSE and their corresponding cluster value K. K starts with 2 and it will increase by 1 in each step. If the graph of line chart shows a drastic decrease in SSE i.e. like an arm, then the "elbow" on the arm is a value to indicate the appropriate number of cluster k in K-means clustering [40].

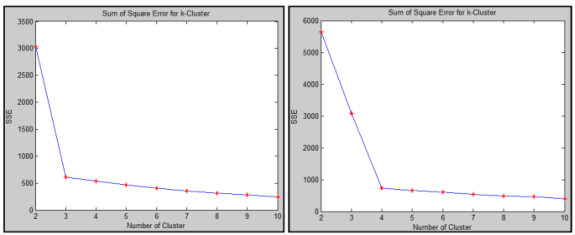

Fig 5. Appropriate number of cluster in the graph by the relationship between SSE and Number of cluster [41].

**For optimal clusters, Algorithm of Elbow method is given**

```
from yellowbrick.cluster import KElbowVisualizer
m = KMeans(random_state=1)
v = KElbowVisualizer(m, k=(2,10))
v.fit(X_numerics)
v.show()
plt.show()
```

The Experimental results for above method is shown in Fig. 6.

Python code is given below for clustering the data after getting optimal number of clusters:

```
m = KMeans(n_clusters=6, random_state=0)
v = SV(model, colors='greenbrick')
v.fit(X_numerics)
v.show()
plt.show()
```

The Experimental results for above method is shown in Fig. 7.

### IV. FIREFLY(FA) AGORITHM

Xin-She Yang given the concept of FA in 2008, as a meta-heuristic optimization based technique.

The FA is based on the attraction behavior of the tropical fireflies and the flashing patterns of their idealized behavior. Since 2010, the FA algorithm has been implemented in different real applications for solving different optimization tasks [42]. The attractiveness of fireflies and variation in light intensity is the important factor of fireflies [43].

Firefly algorithm borrowed 3 ideas from the firefly behavior for a making mathematical model of the algorithms.

1. Because all fireflies are unisex, their attractiveness is not determined by their gender.

2. All Fireflies attract each other on the basis on the basis of their fitness function of the brightness of the light. Fireflies having the worst fitness function move towards better fitness function fireflies.

3. A firefly's brightness and the landscape of the objective function are directly connected.

There are two main tasks we have to define for designing a standard FA, i.e., formulation of light intensity and variation of the attractiveness.

As we know that the intensity of light is inversely proportional to the distance from its source, so it can be approximated as follows. i.e.

$$I = I_0 e^{-\gamma r^2} \quad (11)$$

where $i_0$ denotes the original light intensity at r=0 at the specified source, and γ is the constant coefficient of light absorption.

Now, we can now define the attractiveness of FA as according to the intensity of light proportional which is to be viewing by fireflies adjacent as given below. i.e.

$$\beta = \beta_0 e^{-\lambda r^2} \quad (12)$$

where β 0 denotes light's attraction at r = 0, i.e. its greatest attractiveness. The following is a description of the Cartesian distance rij between firefly i and firefly j:

$$r_{ij} = \|x_i - x_j\|_2 = \sqrt{\sum_{k=1}^{d}(x_{i,k} - x_{j,k})^2} \quad (13)$$

where d is the number of dimensions [44]. The movement of firefly i towards firefly j by their respective brightness is as follows:

$$x_i^{t+1} = x_i^t + \beta e^{\gamma \cdot r_{ij}^2}(x_j^t - x_i^t) + \alpha^t \varepsilon_i^t \quad (14)$$

New solutions of a firefly i depend upon the previous location $x_i$ ones, which is represented by the following equation.

The second part of the above equation is the attraction, whereas in the third term, α is the randomization parameter.

Methods of FA used for clustering is given below with output:

firefly = FireflyAlgorithm(d=d_iris, n=n, range_min=range_min, range_max=range_max,alpha=1.0, beta_max=1.0, gamma=0.5

Output:

firefly {'d': 12, 'n': 150, 'range_min': -5.0, 'range_max': 5.0, 'alpha': 1.0, 'beta_max': 1.0, 'gamma': 0.5}

**FireFly-Algorithm on xclara dataset for automatic clustering**

data_frame = pd.read_csv('../input/testverim/xclara.csv',index_col=False)

cols = [1,2]

data = data_frame[data_frame.columns[cols]]

len_data_points = data.shape[0]

print("Number of data points : {}".format(len_data_points)

print("v1 max : {}, v1 min : {}, v2 max : {}, v2 min : {}".format(v1_max,v1_min,v2_max,v2_min))

Graph of clusters using Firefly on **xclara** data is presented in Fig. 8.

*A. VARIANTS OF FA*

FA divides the entire population into sub-module automatically by attracting the firefly flashes intensity. This is a unique feature of FA compared to other met heuristic algorithms. But original FA stuck into local convergence in most of the dimensions and gave optimal solutions to few dimensions in high dimensional data. Also, it will take high computational time in high dimensional data because all fireflies follow the brighter flashes firefly to their neighborhood [45-46]. Different researchers have proposed different variants of FA to solve the problem of intensification and search diversification in standard FA. Banerjee et al. [47] has given a concept of Propulsion FA by considering three points. 1st used the intensification and diversification meta heuristic search strategy without any parameter tuning, 2nd merging the global best solution from a different dimension in each iteration for the best solution, which acts as a component for a swarm position update. 3rd Manhattan distance measurement has been used in place of Euclidean distance for inter-cluster and intra-cluster distances among data points. The PropFA model has been tested on 18 benchmark functions, 14 additional CEC-2005 functions, and 28 CEC-2013 functions. The analysis shows that the PropFA model's competitiveness is finding the better solutions when it is compared to PSO, RC-Memetic (Real-Coded Memetic), RC-EA (Real-Coded Evolutionary Algorithm), On CEC-2005 benchmark functions, Covariance Adaptation for Matrix Evolution Strategy, and SHADE, Combination of DE and Composite trial vector generation strategies, CEC-2013 benchmark archive i.e. PropFA is a model. This approach is used for manipulating the extension of rapidly spreading oil spills.

FA variants are also frequently used in evolutionary multimodal optimization problems. For solving multimodal optimization problems, the current researcher [39] has given an idea of the FA hybrid method. The researcher divided the FA population into sub-populations using KM. FA was used as a multimodal technique that helps for multiple local optima identification, and K-means enhance the local optima solution. Lot of studies has done, but there are some limitations in the standard FA model due to strict biological laws. Like the position update strategy in FA in equation (3). In this equation, the biological laws of inheritance allow the lesser flashed firefly to approach the position of the higher flashed firefly, limiting the dimensionality and diversity approach because the movement can only occur in the diagonal direction composed of two fireflies.

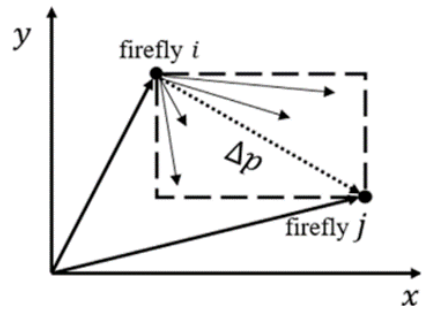

Fig. 9. Firefly movement in 2D, where delta p is the difference of position from fireflies i to j

Instead of searching the entire two-dimensional space, the fireflies move in a diagonal trajectory between two fireflies, as shown in Fig. 1. As a result, the search space is substantially reduced from a two-dimensional region to dashed lines into a one-dimensional vector along the dotted line. In order to mitigate these limitations, we have proposed an Opposition and dimensional-based FA to enhance the exploitation and exploration of search space.

Abbreviations such as IEEE, SI, MKS, CGS, sc, dc, and rms do not have to be defined. Do not use abbreviations in the title or heads unless they are unavoidable.

*B. OPPOSITION AND DIMENSIONAL BASED FA*

Standard FA trap to local optima in high-dimensional data. A meta-heuristic is an excellent algorithm if it effectively balances the intensification and diversification components of the algorithms in the entire search space for getting better performance. Our proposed Opposition-based FA balance the intensification and diversification functions in every generation and fine the global best solution in each dimension, which leads to optimization and getting the global optima.

Algorithm: Opposition and dimensional based FA

Whole Firefly population X= {$X_1$, $X_2$, $X_3$,…...$X_n$} is divided into two groups i.e. $X_i$ and ~$X_i$, where ~$X_i$ is the opposite population of $X_i$ and m<=X<=n

Initialize the population $X_i$={$X_{i1}$,$X_{i2}$,$X_{i3}$,…. $X_{iD}$} and ~$X_{ij}$=$m_j$+$n_j$- $X_{ij}$ and 1<=i<=N and 1<=j<=D

evaluate the fitness function of each firefly by ,

f(Xi)=f($X_{i1}$,$X_{i2}$,$X_{i3}$,…. $X_{iD}$)

Intensity of light of each firefly is equal to f($x_i$)

do

  Loop i=1 to D(dimensions)

    Loop j=1 to N(Firefly)

      Y:=$G_{best}$

      Y:=X$_{(j,i)}$

    End loop

  End loop

  Loop i=1 to N

By applying the supplied formula, you may update the global best in the entire population:

$$r_{ij} = \|x_i - x_j\|_2 = \sqrt{\sum_{k=1}^{d}(x_{i,k}-x_{j,k})^2}$$

attractiveness of firefly is determining by

$$\beta = \beta_0 e^{-\lambda r^2}$$

Update the fireflies with global best

$$x_i^{new} = x_i^{old} + \beta_0 e^{-\gamma r_{ij}^2}(x_i^{old}-G_{bestpos}) + \alpha(rand-\tfrac{1}{2})$$

  End loop

while(t < MaxGenerationFA)

end

## V. PROPOSED CLUSTERING APPROACH BASED ON OPPOSITION MODIFIED FIREFLY ALGORITHM(ODFA) MODEL

In this novel clustering model, there is less probability of centroid initialization sensitivity problem and trapping to local optima problem in comparison to the standard K-means clustering algorithm. For good quality of clusters in K-means, we have to consider some important points: The selection of feasible cluster centroids from the given dataset at the starting and the ability to find global convergence rather than local convergence. As we know, K-means clustering is highly sensitive to the selection of cluster centroids, with parameter k for a number of clusters at the starting. To get good quality clusters, we have to apply K-means clustering many times with random initialization of cluster centroids, which leads to bad convergence results. Our proposed method integrated the K-means with Opposition-based FA because FA used more factors than other nature-inspired optimization algorithms. During the optimization phase, these factors in FA algorithms include particle position, distance, light intensity, and velocity. We have provided more factors in FA that provide better results compared to another optimization algorithm. We divided the clustering task into two steps in our suggested technique; in the 1st step, we employed the firefly algorithm to discover good centroid locations. The objective function in the early steps is to keep the Euclidean distance to a minimum. The number of cores available in the system is the starting point for the method, which starts with the ideal number of clusters. Our Opposition-based FA finds the Gbest solution in every dimension till predefine iteration. In the second step, the initialization of cluster centroids in K-means will be the Gbest of all the fireflies.

## VI. EXPERIMENTAL RESULTS

Experiments were carried out on Iris data sets, the xclara dataset, a customer dataset chosen from the standard data set, UCI, and Kaggle datacamp. We carried out this experiment in Python using the Google Colab online tool. We used principal component analysis to assess the quality of clusters, as shown in Fig. 1. We also used the Silhouette score and the elbow method in our paper to determine the optimal number of clusters in K-means. Figures 5 and 6 show the results. We have also plotted the Silhouette graph in Fig. 7 with their coefficient values and their cluster labels for checking the quality of clusters and their python code is mentioned in the above section III part.

We also wrote Python code for K-means with SilhouetteVisualizer() to produce high-quality clusters. The silhouette graph provided us with the value 6 for K as an input. Figure 8 depicts a Clustered Graph of K-means. We also used the Xclara dataset to implement the FireFly-Algorithm for automatic clustering in Python programming. Section IV of this paper contains the code. The results of the experiments are shown at the end of this section.

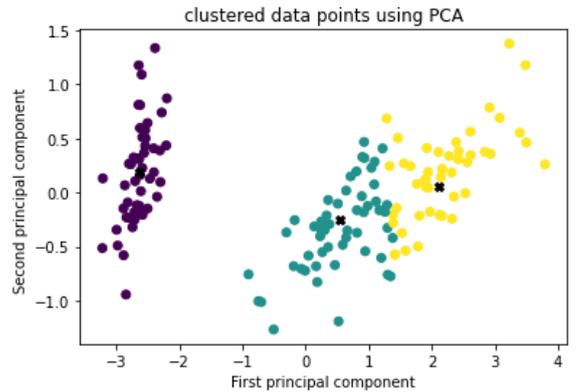

Fig. 1. Clustered data after applying PCA

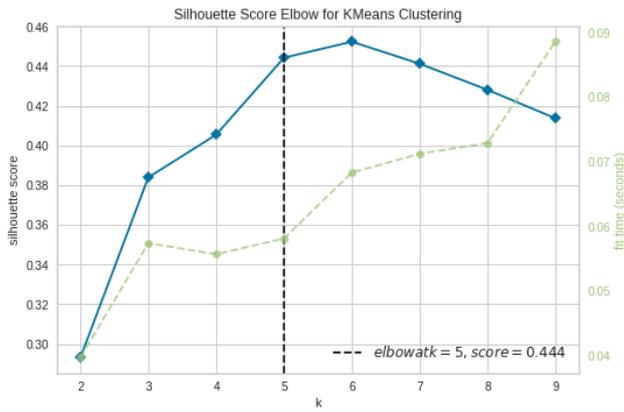

Fig. 3. Silhouette Score and Elbow value for K-means

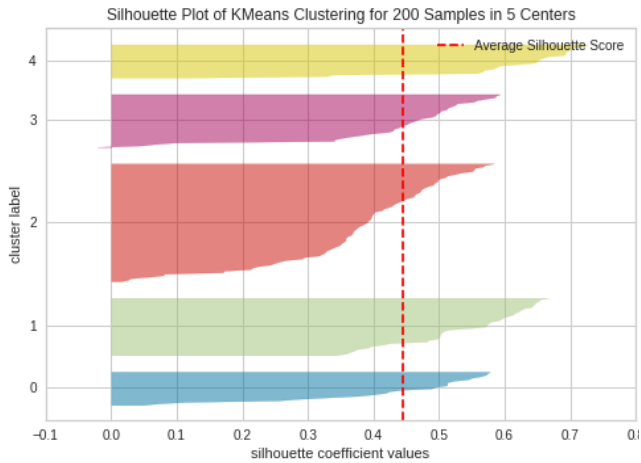

Fig. 4. Quality of Cluster using Silhouette method

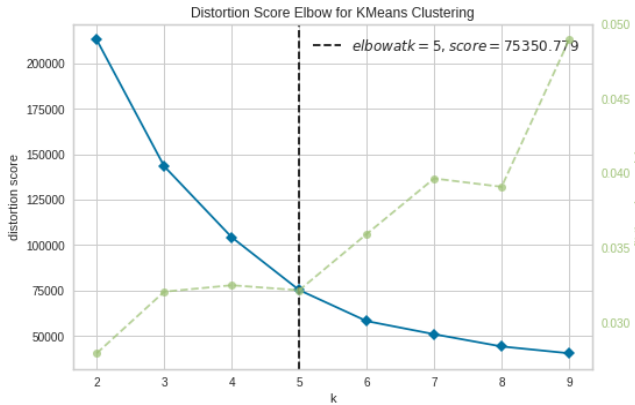

Fig. 6. Elbow method for optimal cluster

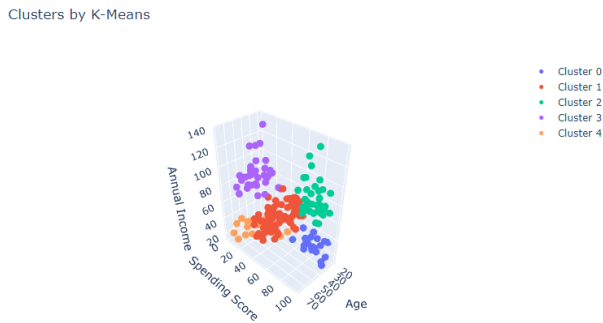

Fig. 7. Data clustered in 6 clusters using K-means

Output of Firefly Algorithm:

Number of data points : 3000

v1 max : 104.3766, v1 min : -22.49599, v2 max : 87.3137, v2 min : -38.7955

**Firefly Matrix:**

[[[77. 6.] [7. 178.] [94. 180.] [[195. 132.][5. 173.] [156. 55.]]

[[114. 151.] [150. 61.] [197. 67.]]

[[69. 132.] [22. 35.] [187. 66.]]

[[191. 54.] [133. 42.] [125. 43.]]]

[0. 605.97229931 582.9588196 355.98673507 530.20899625]

Best firefly matrix:

[[87.63257421 82.49212689]

[-6.37814739 72.15306479]

[-0.848309 79.40971558]]

Initial Centroid Values:

[[87.63257421 82.49212689]

[-6.37814739 72.15306479]

[-0.848309 79.40971558]]

[[69.92418447 -10.11964119]

[9.4780459 10.686052 ]

[ 40.68362784 59.71589274]]

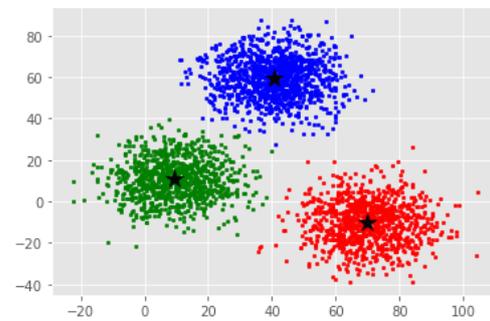

Fig. 8. Data is clustered in 3 clusters using Firefly algorithm

VII. ACKNOWLEDGMENT

VIII. CONCLUSION

Traditional K-means clustering has a lot of challenges, such as being highly reliant on the predefined appropriate number of clusters K. This algorithm also sticks to local optima and it is NP-hard in high-dimensional data. So we have implemented the Silhouette and the Elbow methods with PCA to solve the problem of predefining the appropriate number of clusters K in K-means. We have also hybridized the traditional K-means with a nature-inspired meta-heuristic Firefly optimization algorithm to solve the problem of automatic clustering. The FA divides the entire population into sub-modules automatically by attracting the firefly's flash intensity. This is the unique feature of FA as compared to other swarm-based intelligent algorithms. Due to these unique features, we have explored FA and its variants in our paper. Our proposed opposition-based FA has a higher convergence speed than traditional FA. It also picks up the best from all dimensions. The result of our Silhouette and the Elbow method with PCA and our proposed approach gives the best quality cluster solution in terms of intra-cluster distance and processing CPU time as compare to traditional K-means clustering.